\documentclass[sigconf]{acmart}

\AtBeginDocument{%
  }

\setcopyright{acmlicensed}
\copyrightyear{2024}
\acmYear{2024}
\setcopyright{rightsretained}
\acmConference[MM '24]{Proceedings of the 32nd ACM International Conference on Multimedia}{October 28-November 1, 2024}{Melbourne, VIC, Australia}
\acmBooktitle{Proceedings of the 32nd ACM International Conference on
Multimedia (MM '24), October 28-November 1, 2024, Melbourne, VIC, Australia}
\acmDOI{10.1145/3664647.3681554}
\acmISBN{10.1145/3664647.3681554}

\usepackage{enumitem}
\usepackage{graphicx}
\usepackage[utf8]{inputenc}

\usepackage{amssymb}
\usepackage{booktabs}
\usepackage{diagbox} 
\usepackage{multirow} 
\usepackage{makecell} 
\usepackage{tabularx}
\usepackage{multicol} 
\usepackage{pifont}

\usepackage{orcidlink}
\begin{document}

\title{GRFormer: Grouped Residual Self-Attention for Lightweight Single Image Super-Resolution}

\author{Yuzhen Li}
\orcid{0009-0008-5136-8290}
\affiliation{%
  \institution{Tsinghua University}
  \city{Shenzhen}
  \state{Guangdong}
  \country{China}
}
\email{liyuzhen22@mails.tsinghua.edu.cn}

\author{Zehang Deng}
\authornote{Corresponding Author}
 
\orcid{0009-0000-5469-0762}
\affiliation{%
  \institution{Swinburne University of Technology}
  \city{Melbourne}
  \country{Australia}}
\email{zehangdeng@swin.edu.au}

\author{Yuxin Cao}

\orcid{0009-0002-5766-0846}
\affiliation{%
  \institution{National University of Singapore}
  \country{Singapore}
}
\email{e1374346@u.nus.edu}

\author{Lihua Liu}
\orcid{0009-0006-9536-2005}
\affiliation{%
 \institution{University of International Business and Economics}
 \city{Beijing}
 \country{China}}
 \email{202102027@uibe.edu.cn}

\renewcommand{\shortauthors}{Li et al.}

\begin{abstract}
Previous works have shown that reducing parameter overhead and computations for transformer-based single image super-resolution (SISR) models
(e.g., SwinIR) usually leads to a reduction of performance. In this paper, we present GRFormer, an efficient and lightweight method, which not only reduces the parameter overhead and computations, but also greatly improves performance. The core of GRFormer is Grouped Residual Self-Attention (GRSA), which is specifically oriented towards two fundamental components. Firstly, it introduces a novel grouped residual layer (GRL) to replace the Query, Key, Value (QKV) linear layer in self-attention, aimed at efficiently reducing parameter overhead, computations, and performance loss at the same time. Secondly, it integrates a compact Exponential-Space Relative Position Bias (ES-RPB) as a substitute for the original relative position bias to improve the ability to represent position information while further minimizing the parameter count.
Extensive experimental results demonstrate that GRFormer outperforms state-of-the-art transformer-based methods for $\times$2, $\times$3 and $\times$4 SISR tasks, notably outperforming SOTA by a maximum PSNR of 0.23dB when trained on the DIV2K dataset, while reducing the number of parameter and MACs by about \textbf{60\%} and \textbf{49\% } in only self-attention module respectively. We hope that our simple and effective method that can easily applied to SR models based on window-division self-attention can serve as a useful tool for further research in image super-resolution. The code is available at \url{https://github.com/sisrformer/GRFormer}.
\end{abstract}






\maketitle
\section{Introduction}
\label{sec:intro}
\begin{figure}[tb]
  \centering
  \includegraphics[scale=0.166]{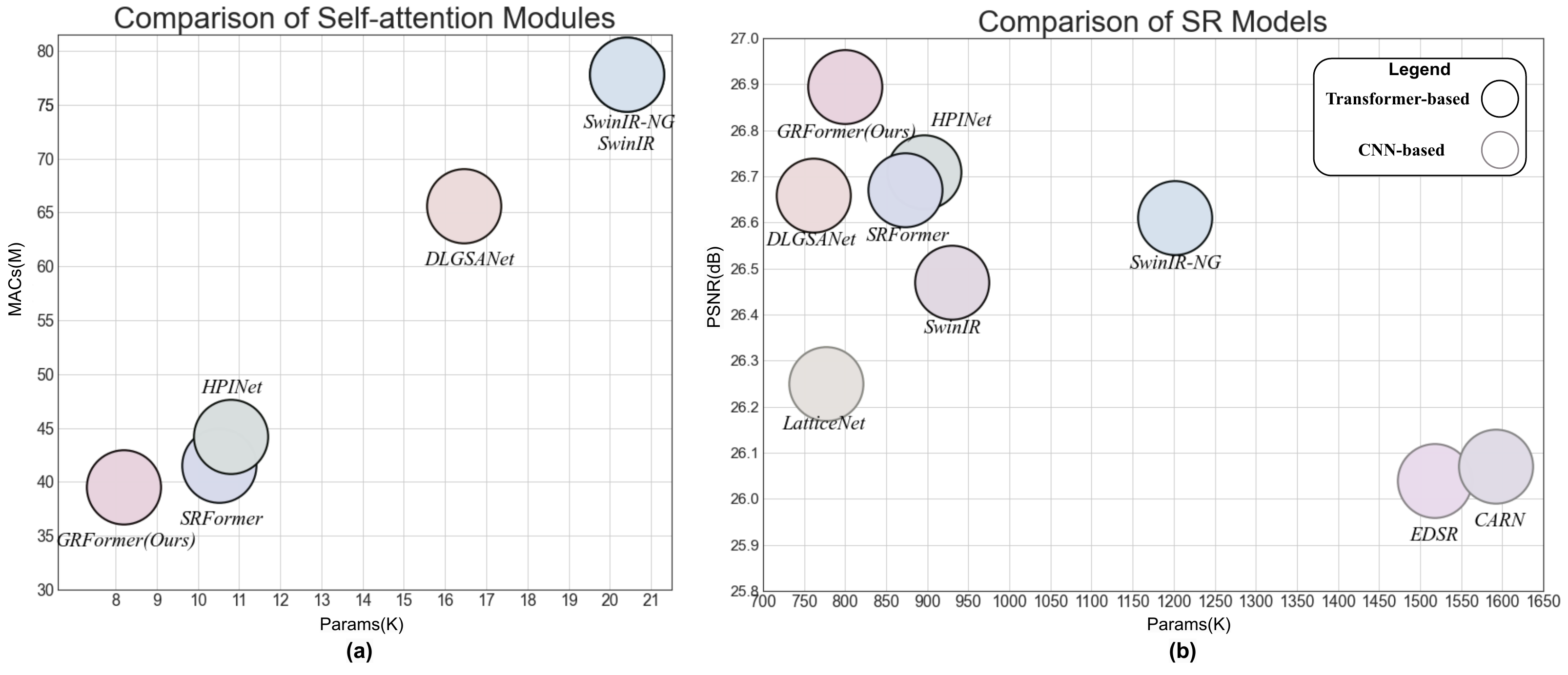}
  \caption{ (a) shows the comparisons of self-attention of recent transformer-based SR models in terms of multiply-accumulate operations (MACs) and parameters. (b) shows SISR comparisons of recent SR models ($\times$4) in terms of PSNR on Urban100, network parameters. Our model (GRFormer) outperforms the SOTA model ($\times$4) by 0.19dB in PSNR score while having comparably low network parameters and MACs.}
  \label{fig:PSNR_Params_MACs_comparison}
\end{figure}

Single Image Super-Resolution (SISR) aims to enhance image resolution by reconstructing a high-resolution image from a low-resolution counterpart.
With the development of CNN-based and transformer-based SR models, one achievement after another has been achieved for single image super-resolution tasks. For whether CNN-based models or transformer-based ones, it is an easy way to improve performance by increasing the number of the network layers and feature dimension, accompanied by the increase of parameters and calculations. A straight question should be: Is it possible to improve the performance while reducing number of the parameters and calculations for transformer-based SR models? 
Motivated by this question, we conduct in-depth research into self-attention and present three sub-questions concerning self-attention:
\vspace{1mm} 
\begin{itemize}[leftmargin=*]
    \item \textbf{RQ1.} Is there any redundancy within self-attention?
    \item \textbf{RQ2.} Can the expressive power of self-attention be further improved?
    \item \textbf{RQ3.} Is there a better alternative to relative position bias for representing the position information?
\end{itemize}
\noindent \textbf{The research into \textbf{RQ1.}} Self-attention \cite{Self-Attention} has achieved great performance in the fields of text, image, and video since it was published in 2017. 
However, despite the effectiveness of self-attention mechanisms, they are often criticized for their extensive parameter count and computational demands. Existing work~\cite{bian2021attention} has observed redundancy in self-attention layers, but its solutions focus mainly on how to reduce the size of the attention window~\cite{zhang2022efficient}. Through empirical analysis, we explore the interaction among the varying parameter counts, MACs, and the performance of the self-attention module, as depicted in Fig.~\ref{fig:PSNR_Params_MACs_comparison}. This analysis reveals significant potential for optimization in both parameters and computational efficiency within the SwinIR's self-attention mechanism. Inspired by these findings, we propose a novel grouping scheme of Q, K, V linear layer, aimed at reducing both the parameter overhead and computational complexity.

\noindent \textbf{The research into \textbf{RQ2.}} 
In the field of single image super-resolution, previous transformer-based approaches ~\cite{SwinIR, SRFormer} have primarily utilized residual learning~\cite{ResidualLearning} at the outer layers of self-attention mechanisms. This methodology has been effective in mitigating network degradation and improving the expressive capabilities of deep networks. Given the success of residual learning in enhancing network performance, an intriguing question arises: could the integration of residual connections within the Query, Key, and Value (QKV) linear layers of self-attention mechanisms further augment their expressive power? Motivated by this consideration, our work explores the incorporation of residual connections directly into the QKV linear layers, aiming to enhance their representational efficacy in deep neural architectures.

\begin{figure}[tb]
  \centering
  \includegraphics[width=0.8\linewidth]{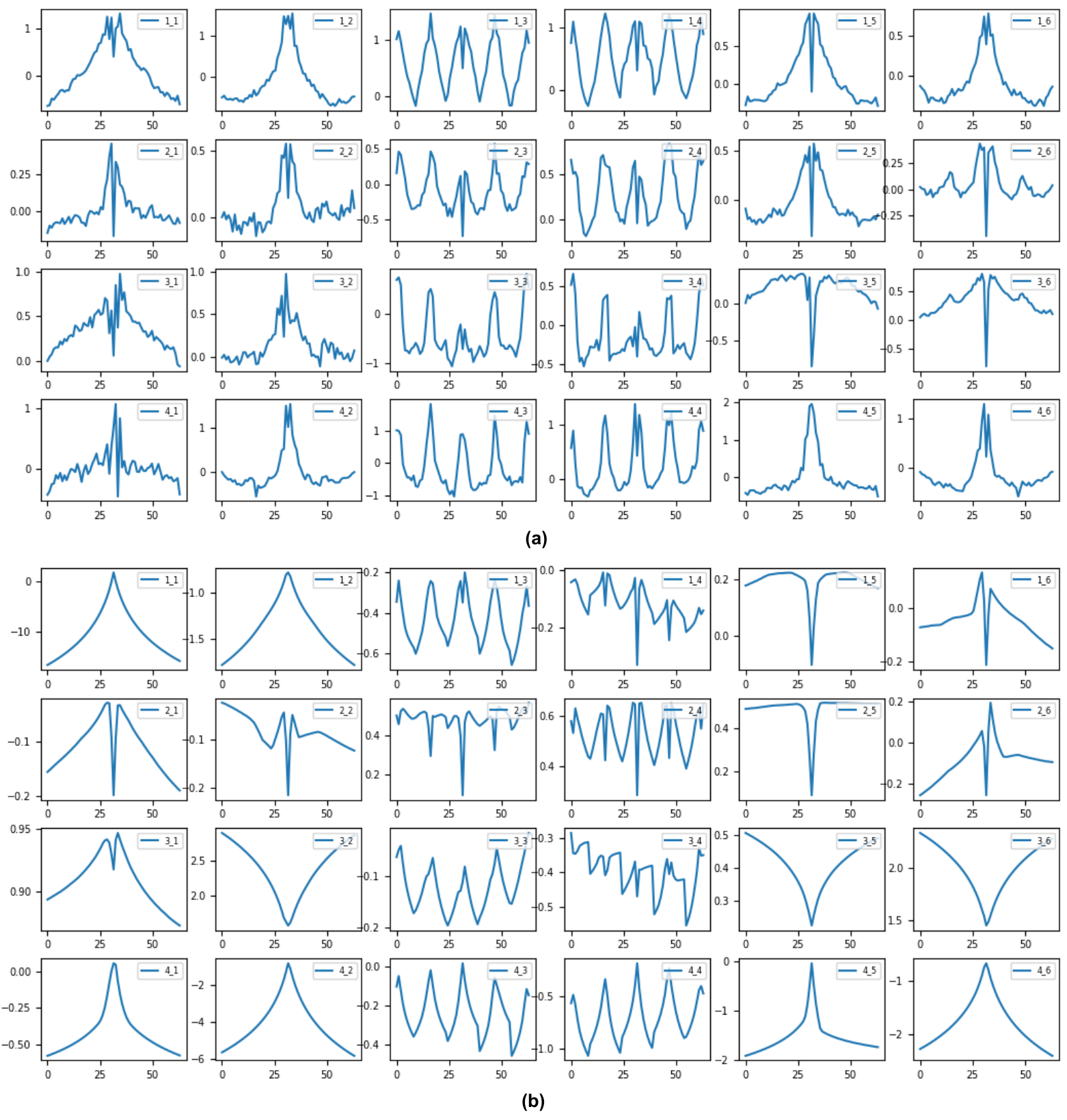}
  \caption{
   Comparison between RPB and ES-RPB. The subfigure (a) and (b) showcase the relative position bias (RPB) from the SwinIR model and an GRFomer model where RPB are replaced to ES-RPB, respectively. The subfigure at $i_{th}$ row and $j_{th}$ column corresponds to the relative position bias of $i_{th}$ GRSAB Group and $j_{th}$ GRSAB in the network. These figures specifically highlight the horizontal evolution of the relative position bias values. The x-axis extends from 0 to 62, while the y-axis corresponds to the data taken at the 7th point on the x-axis from an RPB matrix of size 15$\times$63.
  }
  \label{fig:RPB_comparison}
\end{figure}

\noindent \textbf{The research into \textbf{RQ3.}} When self-attention \cite{Self-Attention} is proposed, it uses position embedding to provide position information of words in text. After self-attention is applied to computer vision, relative position bias (RPB) is found to be more suitable for representing relative position information, which is important for models in computer vision. However, RPB has four fatal flaws: First, if we assume the shape of the window in self-attention is H $\times$ W, the number of parameters occupied by RPB is (2H-1)$\times$(2W-1), which is a huge parameter overhead. Second, the original RPB in SwinIR\cite{SwinIR} sets a position parameter for each position within the window, which is not only the redundancy of the parameters, but also it is easy to be interfered by noises during training. Specifically, many subtle fluctuations can be found in the figures of Fig.~\ref{fig:RPB_comparison} (a). Third, during training, each parameter of RPB is trained independently, ignoring the relative relationship between the weights of different positions. Fourth, RPB fails to clearly reflect intuition: For reconstructing an image, near pixels tend to be more important than far pixels. Motivated by this, we designed an Exponential-Space Relative Position Bias (ES-RPB) to replace RPB.

First and foremost, to solve the \textbf{RQ1}, we propose a grouping scheme for the QKV linear layer. Furthermore, to solve the \textbf{RQ2} and compensate for performance loss from the grouping scheme, we add residuals for the QKV linear layer. Lastly, to solve \textbf{RQ3}, we propose Exponential-Space Relative Position Bias (ES-RPB). 
The three methods above constitute the two fundamental components of Grouped Residual Self-Attention (GRSA). Given the proposed GRSA, we design a lightweight network for SR, termed GRFormer. We evaluate our GRFormer on five widely-used datasets. Benefiting from the proposed GRSA, our GRFormer achieves
significant performance improvements on almost five benchmark datasets. Notably, trained on DIV2K dataset\cite{DIV2K} for $\times$2 SR task, our GRFormer achieves a 33.17 PSNR score on the challenging Urban100 dataset\cite{Urban100}. The result is much higher than recent SwinIR-light\cite{SwinIR}(32.76) and the SOTA lightweight SR model (32.94). This improvement is consistently observed across $\times$3 and $\times$4 tasks. Comprehensive experiments show that GRFormer not only outperforms previous lightweight SISR models~\cite{SwinIR, SRFormer, HPINet}, but also reduces the parameter count by about \textbf{20\%} in total model architecture, compared with SwinIR \cite{SwinIR} (1000k parameters) with the same hyperparameter.
To sum up, our contributions can be summarized as follows:
\vspace{-2mm} 
\begin{itemize}[leftmargin=*]
    \item We propose a novel Grouped Residual Self-Attention (GRSA) for lightweight image super-resolution, which can not only reduces the parameter count but also enhances the performance in an easy-to-understand way for SR tasks. In addition, our proposed GRSA module can seamlessly replace the self-attention module and its variants in other transformer-based SR models, simultaneously reducing the number of parameter by about 60\% and the number of MACs by about 49\% in only self-attention module.
    
    \item Based on GRSA, we construct a novel transformer-based SR network, termed GRFormer. Our GRFormer achieves state-of-the-art performance in lightweight image SR, and outperforms previous lightweight SISR networks by a large margin in most of the five benchmark datasets.
\end{itemize}

\section{Related Works}
\subsection{CNN-Based Image Super-Resolution}
A lot of CNN-based SR models \cite{EDSR, CARN, LatticeNet} have emerged since SRCNN \cite{SRCNN} introduces CNN-based deep learning method for image SR. By removing unnecessary modules in conventional residual networks and using a new multi-scale deep super-resolution system, EDSR \cite{EDSR} achieves significant performance improvement. By implementing a cascading mechanism upon a residual network, CARN \cite{CARN} reduces the amount of computation. To further reduce the amount of computation, LatticeNet \cite{LatticeNet} applies two butterfly structures to combine two residual blocks.

\subsection{Transformer-based Image Super-Resolution}
Since Swin Transformer \cite{SwinTransformer} introduces hierarchical architecture and shifted windowing scheme, the feasibility of transformer application in the field of computer vision has been greatly improved. In order to introduce transformer into the field of image super-resolution, SwinIR\cite{SwinIR} is proposed, which has been baseline model for transformer-based SR models. However, for some lightweight scenarios, the amount of parameters and calculations in SwinIR is still too large and its relative position bias lacks a certain organization. 

\subsection{Relative position bias}
When self-attention is proposed to solve problems in the field of natural language processing (NLP), absolute position embedding is designed to supplement position information of words in text. After self-attention is applied into image super-resolution, relative position bias enjoys more popularity than the absolute position embedding, because relative position bias can provide the relative position information, which is intrinsically more suitable for image super-resolution. However, there are some problems concerning relative position bias that cannot be ignored, such as parameter redundancy, weak ability to resist interference during training and so on. Afterwards, in order to tackle resolution gaps between pre-training of large vision models, SwinIR-v2 ~\cite{SwinIR-V2} proposes a log-spaced continuous position bias method to effectively transfer large-scale models pre-trained using low-resolution images to downstream tasks with high-resolution inputs. Although the log-spaced continuous position bias is designed to solve the problem of resolution difference of pre-trained large-scale models, it is very inspiring for the design of relative position bias of transformer-based SR models.

\section{Method}

\subsection{Overall Architecture}

\begin{figure*}[tb]
  \centering
  \includegraphics[width=0.8\linewidth]{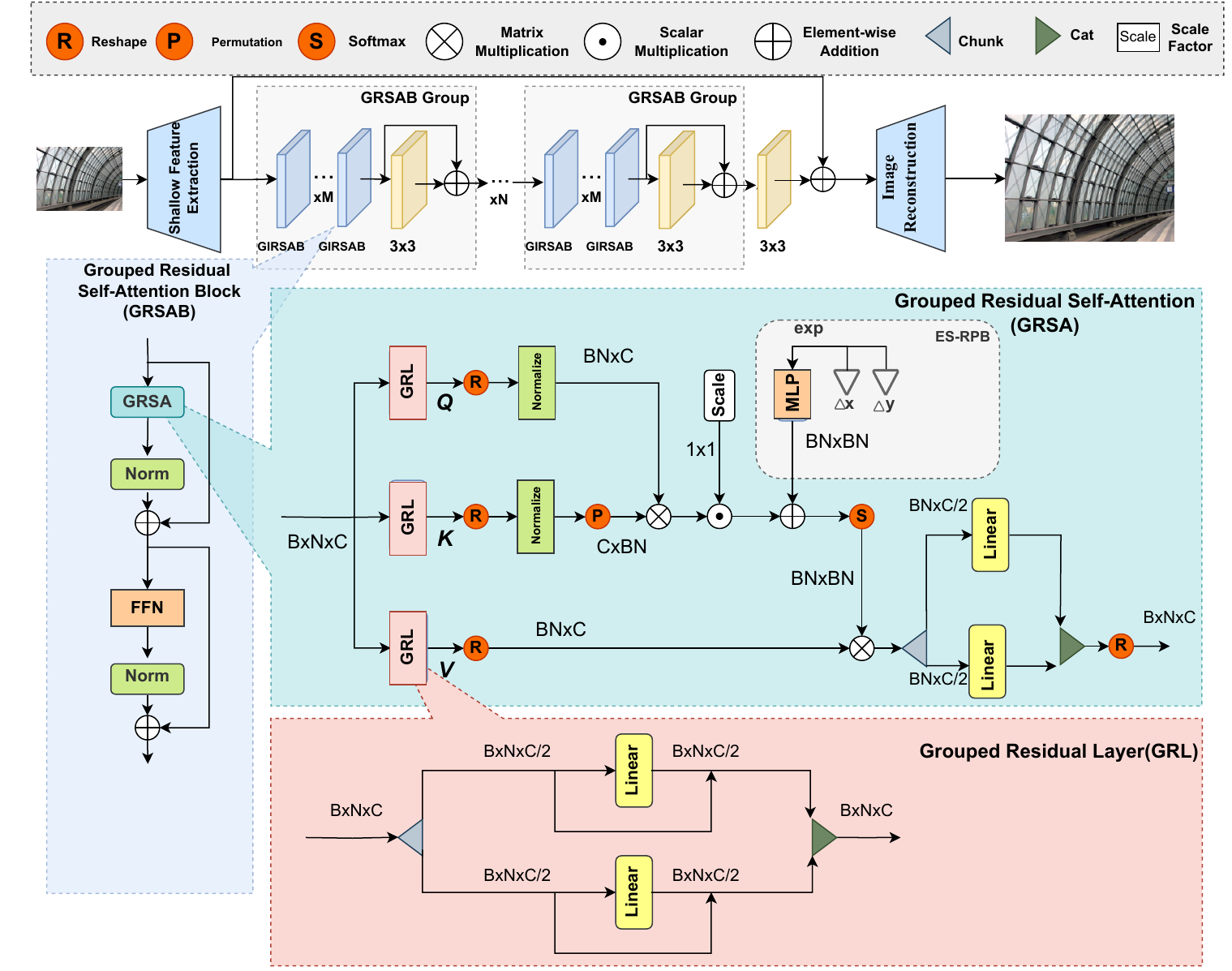}
  \caption{Network architecture of the proposed GRFormer. It mainly consists of a shallow feature extraction module, several grouped residual self-attention block groups (GRSAB Group) to learn deep feature mapping in an efficient and effective way, and a high-resolution image reconstruction module.
}
  \label{fig:GRFormer_overall_network}
\end{figure*}

The overall architecture of our GRFormer is shown in Fig.~\ref{fig:GRFormer_overall_network}, consisting of three parts: shallow feature extraction, deep feature extraction, and image reconstruction. Given the LR input $I_{LR} \in R^{H \times W \times C_{in}}$, we first use a 3$\times$3 convolution $L_{SF}$ to transform the low-resolution image $I_{LR}$ to shallow feature $X_0 \in R^{H \times W \times C}$ as
\begin{equation}
\begin{aligned}
X_0 &= L_{SF}(I_{LR})
\end{aligned}
\label{fl1}
\end{equation}
where $C_{in}$ and C is the channel number of LR input and shallow feature. This convolution layer simply converts the input from image space into high-dimensional feature space. Then, we use N grouped residual self-attention block groups $L_{GRSABG}$ and a 3$\times$3 convolution layer $L_{conv}$ at the end to extract the deep feature $I_{DF} \in R^{H \times W \times C}$. The process can be expressed as
\begin{equation}
\begin{aligned}
X_i &= L_{GRSABG_i}(X_{i-1}), \\
I_{DF} &= L_{conv}(X_N) + X_0 
\end{aligned}
\label{fl2}
\end{equation}
 In GRSAB Group, given $X_i$ as input, we use M grouped residual self-attention block $L_{GRSAB}$ to get $X_{i,M}$. Then we use a 3$\times$3 convolution layer $L_{conv}$ to get $X_{i+1}$. The process can be expressed as
\begin{equation}
\begin{aligned}
X_{i,0} &= X_i, \\
X_{i,j} &= L_{GRSAB_j(X_{i,j-1})}, \\
X_{i+1} &= L_{conv}(X_{i,M}) + X_{i,M} 
\end{aligned}
\label{fl3}
\end{equation}
 Finally, we use a 3$\times$3 convolution layer $L_{conv}$ to get
better feature aggregation, and aggregate shallow and deep features to reconstruct HR image $I_{HR} \in R^{H \times W \times C_{out}}$ as
\begin{equation}
\begin{aligned}
I_{HR} &= L_{shuffle}(L_{conv}(I_{DF}) + I_{SF}), \\
\end{aligned}
\label{fl4}
\end{equation}
where $C_{out}$ is the channel number of the high-resolution image and $L_{shuffle}$ is a PixelShuffle \cite{PixelShuffle} module.

\subsection{Grouped Residual Self-Attention Block}

The \textbf{G}rouped \textbf{R}esidual \textbf{S}elf-Attention \textbf{B}lock (GRSAB) mainly consists of two core components: Grouped Residual Self-Attention (GRSA), Feed Forward Network module (FFN). Given the input of GRSAB as $I_{in} \in R^{H \times W \times C}$, we first use a grouped residual self-attention module $L_{GRSA}$ to learn the global relationships of pixels in a window. After $L_{GRSA}$, we use a LayerNorm module to normalize the feature, because the normalized features can eliminate gradient vanishing and make the training stable. The process can be expressed as
\begin{equation}
\begin{aligned}
I_{GRSA} &= Norm(L_{GRSA}(I_{in})) + I_{in}
\end{aligned}
\label{fl5}
\end{equation}
where $I_{GRSA}$ is the ouput of GRSA module. Then we use a feed forward network to transform the $I_{GRSA}$ to another feature space and a LayerNorm module to normalize the feature as
\begin{equation}
\begin{aligned}
I_{out} &= Norm(L_{FFN}(I_{GRSA})) + I_{GRSA}
\end{aligned}
\label{fl6}
\end{equation}

\subsection{Grouped Residual Self-Attention}

To reduce the number of parameters as well as calculations and enhance the performance, we introduce the Grouped Residual Self-Attention (GRSA), which incorporates two novel and compact components: grouped residual layer (GRL) and exponential-space relative position (ES-RPB). GRL consists of two parts: residuals for QKV linear layer to transform the feature space of self-attention from the linear space to residual space, grouping scheme for QKV linear layer to reduce the parameters and calculations. By reducing the noise interference during training and making self-attention sensitive to pixel distance, ES-RPB improves the expression ability of position information. The proposed GRSA aggregates features of pixels globally in the window. Specifically, given input of GRSA as X $\in R^{H \times W \times C}$, we uses $L_{GRL_Q}$, $L_{GRL_K}$, $L_{GRL_V}$ to get Q, K, V.
\begin{equation}
\begin{aligned}
Q &= L_{GRL_Q}(X), \\
K &= L_{GRL_K}(X), \\
V &= L_{GRL_V}(X)
\end{aligned}
\label{fl6}
\end{equation}
where $L_{GRL_Q}$, $ L_{GRL_K}$, $ L_{GRL_V}$ are the GRL module corresponding to Q, K, V. Then we normalize Q and K and calculate the matrix product of Q and K to get the similarity of Q and K. We multiply $QK^T$ by a trainable self-attention scaling factor $\lambda$, add $B_{ES-RPB}$ and perform a Softmax operation. Next, we calculate the matrix product of the $QK^T$ after Softmax and V. If multi-head self-attention is applied, we will use a grouped linear layer $L_{proj}$ as a projection at the last to map the multi-head to one head. The formulation can be written as:
\begin{equation}
\begin{aligned}
\text{GRSA} = L_{proj}(&\text{Softmax}(\lambda \cdot \text{Normalize}(\textbf{Q}) \\
& * \text{Normalize}(\textbf{K})^\textbf{T} + \textbf{B}_{ES-RPB})\textbf{V})
\end{aligned}
\label{Self-Attention}
\end{equation}

\begin{figure}[tb]
  \centering
  \includegraphics[scale=0.25]{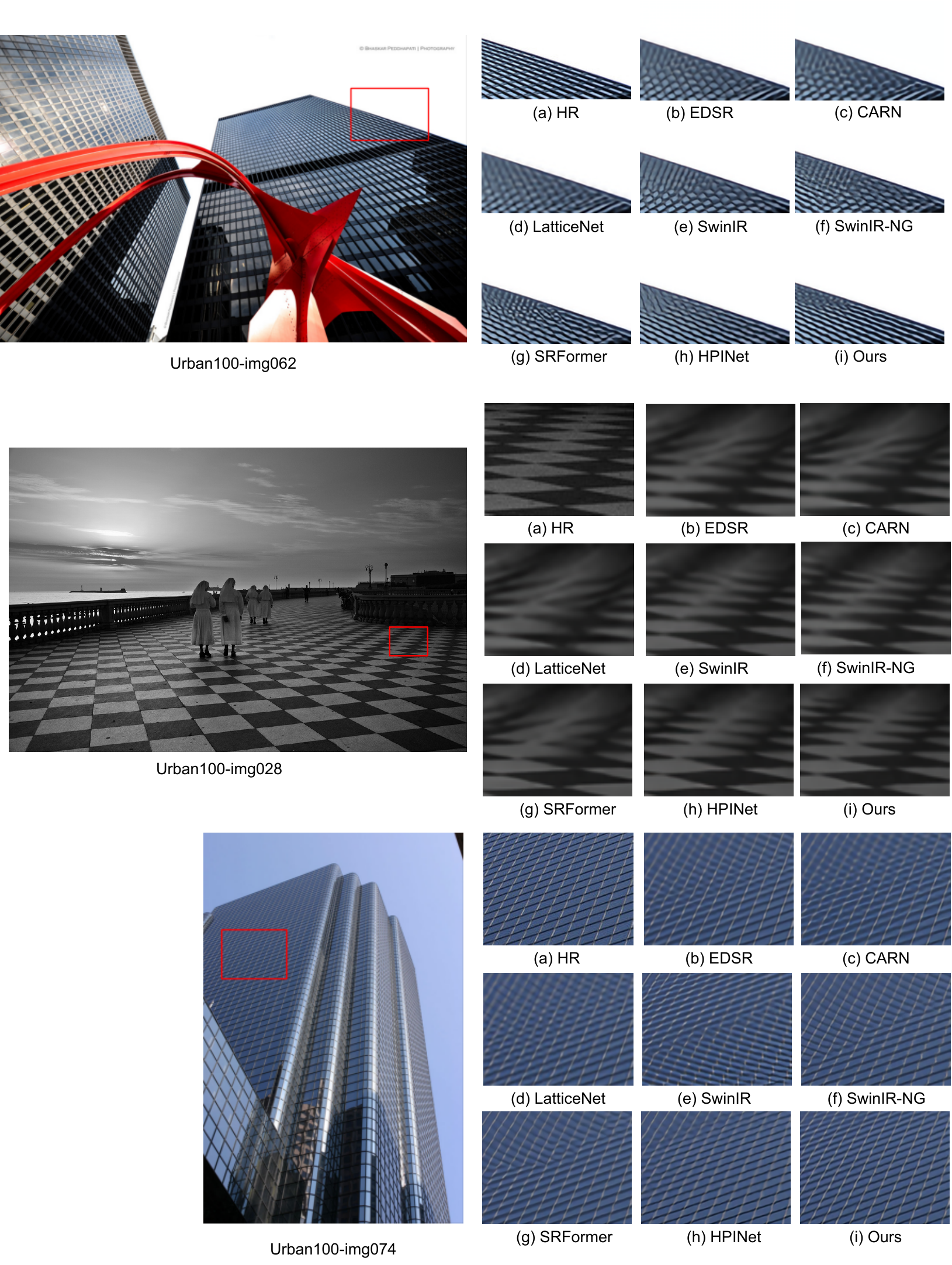}
  \caption{Qualitative comparison with recent state-of-the-art lightweight image SR methods on the ×4 SR task.
}
  \label{fig:expriment_result_comparison_092}
\end{figure}
 
\subsection{Grouped Residual Linear.} 
As a substitute for the QKV linear layer, the Grouped Residual Linear (GRL) is one of the core modules of GRSA, with the objective of reducing the amount of parameters and calculations and basically maintaining the feature learning ability. By integrating concepts of \textit{grouping} and \textit{residuals} into the self-attention mechanism, we enhance its structural efficiency and functional effectiveness.
\begin{itemize}[leftmargin=*]
    \item \textbf{Grouping scheme of QKV linear layer in self-attention.} 
    A lot of works, such as \cite{SRFormer}, show that there is redundancy in generation of Q, K, V and matrix product of Q, K. To reduce the redundancy, we apply the idea of grouping. Given the input as X, we divide X into two equal parts in the channel dimension and then use two independent linear layers to get Q, K, V respectively. Grouping scheme of Q, K, V will not significantly reduce the interaction of pixel features, because the matrix multiplication of Q and K in self-attention will offset these shortcomings to some extent, which is discussed in section 3.6.
    \item \textbf{Residuals of the Q, K, V linear layers.} Residuals allows training to be performed in the residual space, which makes the network find the optimal solution in the residual space.  So we add it into the QKV linear layer to transform the training space of QKV linear layer from linear space to residual space in order to enhance the feature learning ability of QKV linear layer.
\end{itemize}
Specifically, we assume the input of GRL module is X $\in R^{H \times W \times C}$. We first divide X in channel dimension into two parts $X_{in_1}$, $X_{in_2}$. Then, by using residuals, each uses a linear layer to get $X_{out_1}$, $X_{out_2}$. Finally, we merge $X_{out_1}$, $X_{out_2}$ in channel dimension to get the output $X_{out}$ at the last as
\begin{equation}
\begin{aligned}
X_{in_1}, X_{in_2} &= X, \\
X_{out_1} &= L_{Linear_1}(X_{in_1}) + X_{in_1}, \\
X_{out_2} &= L_{Linear_2}(X_{in_2}) + X_{in_2}, \\
X_{out} &= X_{out_1}, X_{out_2}, \\
\end{aligned}
\label{fl1}
\end{equation}
where $L_{Linear_1}$ and $L_{Linear_2}$ are linear layers, $X_{in_1}$, $X_{in_2}$, $X_{out_1}$, $X_{out_2}$ $\in R^{H \times W \times \frac{C}{2}}$.

\subsection{Exponential-Space Relative Position Bias} 
 To solve the four disadvantages of original RPB mentioned in \textbf{RQ3}, we propose the Exponential-Space Relative Position Bias (ES-RPB). We design a exponential mapping for original absolute position coordinates to forcibly add pixel distance sensitive rules to original RPB, which makes it give more weight to nearby pixels. What's more, we use a tiny multilayer perceptron (MLP) to obtain the mapping of all absolute position coordinates, which reduces the impact of noise during training and makes it easier to be trained. Specifically, we transform the abscissa $\Delta$X and the ordinate $\Delta$Y in absolute position coordinate matrix from linear space to exponential space, and then we use a tiny MLP to get $B_{ES-RPB}$
 
 \begin{equation}
 \begin{aligned}
 \Delta \widehat{X} &= sign(\Delta X) * (1 - \exp(-|\alpha * \Delta X|)), \\
 \Delta \widehat{Y} &= sign(\Delta Y) *  (1 - \exp(-|\beta *\Delta Y|)),  \\
 B_{ES-RPB} &= MLP(\Delta \widehat{X}, \Delta \widehat{Y})
 \end{aligned}
 \label{eq:ES-RPB}
 \end{equation}
where $\alpha$ and $\beta$ is trainable distance-sensitive factors to control the sensitivity to distances between reference pixel and the others in the same window and the symbol * represents the multiplication of each position in the matrix. MLP consists of two linear layers and an activation layer sandwiched between them.

\begin{table*}[!t]
 \renewcommand\arraystretch{1.}
 \begin{center}
 \caption{Quantitative evaluations of the lightweight GRFormer against state-of-the-art methods on commonly used benchmark datasets. Best and second best results are marked in \textcolor{red}{red} and \textcolor{blue}{blue} colors. \#Params means the number of the network parameters. \#MACs denotes the number of the MACs which are calculated on images with an upscaled spatial resolution of 1280 $\times$ 720 pixels. \#Weighted-Avg means the  weighted average PSNR and SSIM on five benchmark datasets. \#Weighted-Avg = $\sum_{i=1}^5$ $M_i$ $\times$ $Score_i$, where M is the number of images in the dataset and Score is the corresponding PSNR or SSIM score.}
 \label{tab:QuantitativeComparison}
 \vspace{-3mm}
  \resizebox{0.95\textwidth}{!}{
  \small
   \begin{tabular}{| c | l | c | c | c | l | l | l | l | l | l |}
    \hline
    Scale & Method & Year & \#Params(/K) & \#MACs(/G) & Set5 & Set14 & B100 & Urban100 & Manga109 & \#Weighted-Avg\\
    \hline
    \multirow{6}*{$\times 2$}
      & EDSR-baseline\cite{EDSR}  &  CVPRW2017&1370      & 316.3    & 37.99/0.9604 & 33.57/0.9175 & 32.16/0.8994 & 31.98/0.9272 & 38.54/0.9769 & 34.37/0.9353 \\
      & CARN~\cite{CARN}   & ECCV2018& 1592        & 222.8     &  37.76/0.9590  & 33.52/0.9166   &  32.09/0.8978  &   31.92/0.9256  &  38.36/0.9765 & 34.27/0.9342 \\
      & LatticeNet~\cite{LatticeNet} & ECCV2020 & 756       & 169.7    & 38.15/0.9610 & 33.78/0.9193 & 32.25/0.9005 & 32.43/0.9302 & -/-& -/-\\
      & SwinIR-light~\cite{SwinIR} & ICCV2021  & 910       & 207.5    & 38.14/0.9611 & 33.86/0.9206 & 32.31/0.9012 & 32.76/0.9340 & 39.12/\textcolor{blue}{0.9783} & 34.87/0.9386\\
      & SwinIR-NG~\cite{SwinIR-NG} & CVPR2023 & 1181& 210.7& 38.17/0.9612&\textcolor{blue}{33.94}/0.9205&32.31/0.9013& 32.78/0.9340&39.20/0.9781 & 34.90/0.9385\\
      & DLGSANet-light~\cite{DLGSANet} & ICCV2023 & 745& 169.4 & 38.20/0.9612 & 33.89/0.9203 & 32.30/0.9012 & \textcolor{blue}{32.94}/\textcolor{blue}{0.9355} & \textcolor{blue}{39.29}/0.9780& \textcolor{blue}{34.98}/0.9389 \\
      & SRFormer-light ~\cite{SRFormer}  & ICCV2023 &   853    &   198.6   &  \textcolor{red}{38.23}/\textcolor{blue}{0.9613} & \textcolor{blue}{33.94}/\textcolor{blue}{0.9209} & \textcolor{red}{32.36}/\textcolor{red}{0.9019} & 32.91/0.9353 &39.28/\textcolor{red}{0.9785} & \textcolor{blue}{34.98}/\textcolor{blue}{0.9393}\\
      & HPINet-M ~\cite{HPINet}  & AAAI2023 &  783     &  213.1    &   38.12/- &  \textcolor{blue}{33.94}/- &  32.31/- & 32.85/- &  39.08/- & 34.88/-\\
      & \textbf{GRFormer (Ours)}  & - &  781     &    198.4 & \textcolor{blue}{38.22}/\textcolor{red}{0.9614} & \textcolor{red}{34.01}/\textcolor{red}{0.9214} & \textcolor{blue}{32.35}/\textcolor{blue}{0.9018} & \textcolor{red}{33.17}/\textcolor{red}{0.9375} & \textcolor{red}{39.30}/\textcolor{red}{0.9785} & \textcolor{red}{35.07}/\textcolor{red}{0.9399}\\

    \hline

    \multirow{6}*{$\times 3$}
      & EDSR-baseline~\cite{EDSR}  &CVPRW2017& 1555      & 160.1    & 34.37/0.9270 & 30.28/0.8417 & 29.09/0.8052 & 28.15/0.8527 & 33.45/0.9439 & 30.38/0.8692\\
      & CARN~\cite{CARN}  & ECCV2018&1592  & 118.6   &  34.29/0.9255  & 30.29/0.8407  & 29.06/0.8034 &  28.06/0.8493 & 33.50/0.9440 & 30.36/0.8676\\
      & LatticeNet~\cite{LatticeNet} &ECCV2020& 765       & 76.2     & 34.53/0.9281 & 30.39/0.8424 & 29.15/0.8059 & 28.33/0.8538 & -/-& -/-\\
      & SwinIR-light~\cite{SwinIR}   &  ICCV2021&918       & 94.2     & 34.62/0.9289 & 30.54/0.8463 & 29.20/0.8082 & 28.66/0.8624 & 33.98/0.9478 & 30.76/0.8746\\
      & SwinIR-NG~\cite{SwinIR-NG} & CVPR2023 & 1190 & 95.67 & 34.64/0.9293&30.58/\textcolor{blue}{0.8471}&29.24/0.8090&28.75/0.8639& \textcolor{blue}{34.22}/0.9488& 30.89/0.8757\\
      & DLGSANet-light~\cite{DLGSANet}&ICCV2023& 752 & 75.8     & \textcolor{red}{34.70}/\textcolor{blue}{0.9295} & 30.58/0.8465 & 29.24/0.8089 & 28.83/0.8653 & 34.16/0.9483 & 30.89/0.8759\\
        & SRFormer-light ~\cite{SRFormer}  & ICCV2023& 861       &  88.3   & \textcolor{blue}{34.67}/\textcolor{red}{0.9296}  & 30.57/0.8469 & \textcolor{blue}{29.26}/\textcolor{blue}{0.8099} & 28.81/\textcolor{blue}{0.8655} & 34.19/\textcolor{blue}{0.9489} & 30.90/\textcolor{blue}{0.8764}\\
        & HPINet-M ~\cite{HPINet}  & AAAI2023& 924       &  110.6   &  \textcolor{red}{34.70}/-  &  \textcolor{blue}{30.63}/- & \textcolor{blue}{29.26}/- & \textcolor{blue}{28.93}/- & 34.21/- & \textcolor{blue}{30.95}/-\\
      & \textbf{GRFormer (Ours)}   & - & 789&93.5& \textcolor{blue}{34.67}/0.9293& \textcolor{red}{30.64}/\textcolor{red}{0.8481} & \textcolor{red}{29.27}/\textcolor{red}{0.8100} & \textcolor{red}{29.07}/\textcolor{red}{0.8702} & \textcolor{red}{34.35}/\textcolor{red}{0.9494} & \textcolor{red}{31.04}/\textcolor{red}{0.8781}\\
      
    \hline

    \multirow{6}*{$\times 4$}
      & EDSR-baseline~\cite{EDSR}  &CVPRW2017& 1518      & 114.2    & 32.09/0.8938 & 28.58/0.7813 & 27.57/0.7357 & 26.04/0.7849 & 30.35/0.9067 & 28.14/0.8119\\
      & CARN~\cite{CARN} &ECCV2018& 1592  & 90.88 &  32.13/0.8937 & 28.60/0.7806 &  27.58/0.7349 &  26.07/0.7837 & 30.47/0.9084 & 28.19/0.8118\\
      & LatticeNet~\cite{LatticeNet}&ECCV2020& 777       & 43.6     &  32.30/0.8962 &  28.68/0.7830 & 27.62/0.7367 & 26.25/0.7873 & -/-& -/-\\
      & SwinIR-light~\cite{SwinIR}&ICCV2021& 930       & 54.18     & 32.44/0.8976 & 28.77/0.7858 & 27.69/0.7406 & 26.47/0.7980 & 30.92/0.9151 & 28.51/0.8204\\
        & SwinIR-NG~\cite{SwinIR-NG} & CVPR2023 & 1201       & 55      & 32.44/0.8980&28.83/0.7870&\textcolor{blue}{27.73}/0.7418& 26.61/0.8010&31.09/0.9161 & 28.62/0.8221\\
      & DLGSANet-light ~\cite{DLGSANet}  & ICCV2023&761& 43.2  & 32.54/\textcolor{red}{0.8993} & 28.84/0.7871 & \textcolor{blue}{27.73}/0.7415 & 26.66/\textcolor{blue}{0.8033} & 31.13/0.9161 & 28.65/0.8227\\
      & SRFormer-light ~\cite{SRFormer}  &ICCV2023& 873       & 53     & 32.51/\textcolor{blue}{0.8988}  & 28.82/\textcolor{blue}{0.7872} & \textcolor{blue}{27.73}/\textcolor{blue}{0.7422} & 26.67/0.8032 & 31.17/\textcolor{blue}{0.9165} & 28.67/\textcolor{blue}{0.8230}\\
      & HPINet-M ~\cite{HPINet}  & AAAI2023& 896       &  81.1   &  \textcolor{red}{32.60}/-   &  \textcolor{blue}{28.87}/- & \textcolor{blue}{27.73}/- &  \textcolor{blue}{26.71}/- & \textcolor{blue}{31.19}/- & \textcolor{blue}{28.69}/-\\
      & \textbf{GRFormer (Ours)}   & -&  800     &   50.8   & \textcolor{blue}{32.58}/\textcolor{red}{0.8994} & \textcolor{red}{28.88}/\textcolor{red}{0.7886} & \textcolor{red}{27.75}/\textcolor{red}{0.7431} & \textcolor{red}{26.90}/\textcolor{red}{0.8097} & \textcolor{red}{31.31}/\textcolor{red}{0.9183} & \textcolor{red}{28.80}/\textcolor{red}{0.8260}\\
    \hline

  \end{tabular}}
 \end{center}
 \vspace{-6mm}
\end{table*}

\subsection{Explanation of the effectiveness of GRL}
GRL consists of two parts: a grouping scheme and a residual structure. Grouping scheme is proposed to efficiently reduce the parameters and calculations without severe performance degradation, and residual structure makes the network find the optimal solution in the residual space, thereby reducing the difficulty of feature learning. The explanation is as follows:

\begin{itemize}[leftmargin=*]
    \item \textbf{Explanation of the effectiveness of grouping scheme.}
    First, we analyse the effectiveness in reduction of parameters and MACs.
    We assume that the number of input features is N. Then, both the parameters and the MACs occupied by the linear layer are $N^2$. But, if we group the input features into halves and use two linear layers with the number of input features of $\frac{N}{2}$ to process the two parts of input features, respectively, both parameters and MACs occupied by the two linear layers are only $\frac{N^2}{2}$. Obviously, the grouping scheme can halve the number of parameters and MACs. 
    
    Second, we analyse the effectiveness analysis in preventing severe performance degradation.
     A possible concern about the grouping scheme is that grouping features will result in a lack of aggregation of the two groups of features, leading to performance degradation. We will analyze below that, at least for Q, K matrices, this worry is unnecessary. As shown in the self-attention formula: $\text{Attention}(Q, K, V) = \text{Softmax}\left(\frac{QK^T}{\sqrt{d_k}}\right)V$, Q and K matrices are only used to perform matrix product and get $QK^T$ matrix. During the matrix product of Q and K, even if we employ the grouping scheme to divide the input features into two groups and perform feature aggregation individually, the subsequent matrix product of Q and K will deeply aggregate the two groups of features. The process can be shown as follows.
     
     Specifically, given input of self-attention as X $\in$ $R^{N \times C}$, where N is the number of pixels in a window and C is number of input features, we cut up X into 2$\times$2 blocks as 
     \begin{equation}
     \begin{aligned}
     \begin{pmatrix} X_{11} & X_{12} \\ X_{21} & X_{22} \end{pmatrix} &= X, \\
     \end{aligned}
     \label{eq:block matrix X}
     \end{equation}
     where $X_{11}$, $X_{12}$, $X_{21}$, $X_{22}$ $\in$ $R^{\frac{N}{2} \times \frac{C}{2}}$.
     By packing parameter matrices of two grouped linear layers with input features of $\frac{C}{2}$ into one parameter matrix, we can assume the parameter matrices of Q, K projections as $\begin{pmatrix} M_{Q_1} \\ M_{Q_2}\end{pmatrix}$, $\begin{pmatrix} M_{K_1} \\ M_{K_2}\end{pmatrix}$ respectively, where $M_{Q_1}$, $M_{Q_2}$, $M_{K_1}$, $M_{K_2}$ $\in$ $R^{\frac{C}{2} \times \frac{C}{2}}$. Then we can perform matrix product of X and the parameter matrices of Q, K projections respectively as
     \begin{equation}
     \begin{aligned}
     Q,K  &= \begin{pmatrix} X_{11} & X_{12} \\ X_{21} & X_{22} \end{pmatrix} * \begin{pmatrix} M_{Q_1 \vee K_1} \\ M_{Q_2 \vee K_2}\end{pmatrix} \\
     &= \begin{pmatrix} X_{11} * M_{Q_1 \vee K_1} & X_{12} * M_{Q_2 \vee K_2} \\ X_{21} * M_{Q_1 \vee K_1} & X_{22} * M_{Q_2 \vee K_2} \end{pmatrix}, 
     \end{aligned}
     \label{eq:cal_QKV}
     \end{equation}
     where $\vee$ represents the logical symbol "or". Then, we perform the matrix product of Q and K as follows:
\begin{equation}
\begin{aligned}
QK^T &= \begin{pmatrix}
O_{11} & O_{12} \\
O_{21} & O_{22}
\end{pmatrix},
\end{aligned}
\label{oo}
\end{equation}
then, each element in matrix of $QK^T$ is calculated:
\begin{equation}
\begin{aligned}
O_{11} &= X_{11} M_{Q_1} {M_{K_1}}^T {X_{11}}^T + X_{12} M_{Q_2}{M_{K_2}}^T{X_{12}}^T,\\
O_{12} &= X_{11} M_{Q_1} {M_{K_1}}^T {X_{21}}^T + X_{12} M_{Q_2} {M_{K_2}}^T {X_{22}}^T, \\
O_{21} &= X_{21} M_{Q_1} {M_{K_1}}^T {X_{11}}^T + X_{22} M_{Q_2} {M_{K_2}}^T {X_{12}}^T, \\
O_{22} &= X_{21} M_{Q_1} {M_{K_1}}^T {X_{21}}^T + X_{22} M_{Q_2} {M_{K_2}}^T {X_{22}}^T
\end{aligned}
\label{ij}
\end{equation}
    
     It can be seen from Equation~\ref{ij}, in the matrix product of Q and K of the self-attention mechanism, both $X_{11}$ and $X_{12}$ are used to perform a series of matrix products to get $O_{11}$, $O_{12}$, and so do $X_{21}$ and $X_{22}$ to get $O_{21}$, $O_{22}$, which aggregates two groups of features. Therefore, the matrix product of Q and K will offset the weak aggregation of input features brought by the use of grouping scheme. In other words, our grouping scheme won't lead to lack of aggregation of the two groups of features and thereby won't lead to severe performance degradation.
     
    \item \textbf{Explanation of the effectiveness of residual structure for the QKV linear layer.} As shown in Formula \ref{Self-Attention}, self-attention involves multiple matrix products, which makes it difficult for the network to learn the optimal parameters. Given the input of self-attention as X and the optimal Q, K, V as $\widehat{Q}$, $\widehat{K}$, $\widehat{V}$, if we add the residual structure for QKV linear layer, the network only needs to optimize $\widehat{Q}$-X, $\widehat{K}$ - X, $\widehat{V}$ - X instead of $\widehat{Q}$, $\widehat{K}$, $\widehat{V}$. The residual structure transforms the learning space of QKV linear layer from linear space to residual space, enhancing the feature learning ability of QKV linear layer.
\end{itemize}

\subsection{Explanation of the effectiveness of ES-RPB}
We mainly analyze the effectiveness of ES-RPB from the perspective of parameter reduction and performance improvement for ES-RPB.
\begin{itemize}[leftmargin=*]
    \item \textbf{Explanation of parameter reduction for ES-RPB.} We assume that the size of the window of self-attention is W$\times$H and the self-attention head is 1. The number of parameters occupied by the original RPB is (2$\times$W-1)$\times$(2$\times$H-1). Specifically, the window size is usually set to 16$\times$16, so the parameter count is 961 in this case. In contrast, if we suppose that the number of features of the hidden layer in MLP is $C_{hidden}$, the amount of parameters occupied by ES-RPB is 3$C_{hidden}$. Specifically, $C_{hidden}$ is set to 128 in our GRFormer, so the number of parameters occupied by ES-RPB in GRFormer is 384, which is much less than 961. What has to be noted is that, when the height and width of window grow simultaneously at a linear rate, the parameters occupied by original RPB will grow squarely, while the parameters occupied by ES-RPB will not grow. 

    \item \textbf{Explanation of performance improvement for ES-RPB.} The ES-RPB mechanism within GRFormer improves performance through three key strategies. Firstly, instead of training positional parameters directly, which can be noise-sensitive, we utilize a tiny MLP to get the $B_{ES-RPB}$. This approach minimizes the noise impact on positional parameters during training. Secondly, the mechanism enhances interaction of parameters representing relative position information by training them through this tiny MLP rather than in isolation. Thirdly, ES-RPB introduces a distance-sensitive design. It employs an exponential function to map the absolute positional coordinates (\(\Delta X, \Delta Y\)) from linear space to exponential space, resulting in \(\Delta \widehat{X}\) and \(\Delta \widehat{Y}\). This transformation ensures that positions closer to the reference pixel exhibit more significant changes, aligning with the principle that nearer pixels should attract more attention. 
\end{itemize}

\section{Experiments}
In this section, we conduct experiments on the lightweight image SR tasks, compare our GRFormer with existing state-of-the-art methods, and do ablation analysis of the proposed method.

\subsection{Experimental Setup}

\textbf{Datasets and Evaluation}. 
For training, we use DIV2K (Agustsson and Timofte 2017), the same as the comparison models, to train our GRFormer. It includes 800 training images and 100 validation images, mainly concerning human, animals, plants, buildings, etc. For testing, we use five public SR benchmark datasets: Set5 (Bevilacqua et al.
2012), Set14 (Zeyde, Elad, and Protter 2010), B100 (Martin
et al. 2001), Urban100 (Huang, Singh, and Ahuja 2015) and
Manga109 (Matsui et al. 2017) to evaluate model.
The experimental results are evaluated in terms of two objective criteria: peak signal-to-noise ratio (PSNR) and structural similarity index (SSIM), which are both calculated on the Y channel from the YCbCr space. 

\begin{table*}[!t]
  \caption{Effect of the GRFormer on SISR. \#SA-Params and \#SA-MACs mean the parameters and MACs in our GRSA respectively. \#Params and \#MACs mean the parameters and MACs in our GRFormer. The ablation experiments are trained on DF2K for $\times$4 SR task and tested on benchmark datasets (Set5, Set14, B100, Urban100, Manga109) to get PSNR and SSIM.}
  \vspace{-3mm}
    \centering
  \label{tab: AblationAnalysis}
\footnotesize
\resizebox{1.0\textwidth}{!}{
 \centering
 \begin{tabular}{c|c|c|c|c|c|c|c|c|c|c|c|c}
    \toprule
     \multirow{2}{*}{Method} & \multicolumn{2}{|c|}{GRL} & \multirow{2}{*}{ES-RPB} & \multirow{2}{*}{\#SA-Params} & \multirow{2}{*}{\#Params} & \multirow{2}{*}{\#SA-MACs} & \multirow{2}{*}{\#MACs}  & \multirow{2}{*}{Set5} & \multirow{2}{*}{Set14} & \multirow{2}{*}{B100} & \multirow{2}{*}{Urban100} & \multirow{2}{*}{Manga109}\\
     \cline{2-3}
     & Group & Residuals &  & & & & & & & & &\\
    \hline
        \textcircled{1}&\ding{51} & \ding{51} &  \ding{51} & 8.2K & 810K & 39.5M & 50.8G  & 32.59/0.8999 & 28.92/0.7890&27.77/0.7435 & 26.97/0.8110 & 31.47/0.9195\\
         \textcircled{2}&\ding{51} & \ding{51} &  & 10.3K & 850K & 38.9M & 50.8G  & 32.46/0.8980 & 28.81/0.7865  & 27.72/0.7414& 26.53/0.7998 & 31.07/0.9150\\
        \textcircled{3}& & &  \ding{51} & 15.4K & 973K & 69M & 61.4G  & 32.60/0.9000 & 28.93/0.7892 &27.77/0.7437 & 27.00/0.8118 & 31.48/0.9197\\
        \textcircled{4} & & \ding{51} & \ding{51} & 15.4K & 973K & 69M & 61.4G & 32.63/0.9002 & 28.95/0.7897 & 27.78/0.7441 & 27.07/0.8139 & 31.56/0.9206 \\
        \textcircled{5}& \ding{51} & &  \ding{51} & 8.2K & 810K & 39.5M & 50.8G  & 32.58/0.8996 & 28.91/0.7888 & 27.76/0.7434 & 26.97/0.8116 & 31.43/0.9191\\
        \textcircled{6}&  & &  & 20.4K & 1023K & 77.8M & 64.8G  & 32.53/0.8988 & 28.85/0.7874 2 & 27.74/0.7424 & 26.64/0.8030 & 31.17/0.9165\\
 \bottomrule
  \end{tabular}
}
\vspace{-1mm}
\end{table*}

\begin{table*}[!t]
  \caption{Quantitative evaluations for $\times$4 task. GRFormer-8$\times$32, GRFormer-16$\times$16 means GRFormer that uses window size of 8$\times$32, 16$\times$16 respectively.}
  \vspace{-3mm}
    \centering
  \label{tab: WindowSizeComparison}
\footnotesize
\resizebox{0.8\textwidth}{!}{
 \centering
 \begin{tabular}{c|c|c|c|c|c|c|c|c}
    \toprule
     Method & \#Params(/K) & \#MACs(/G)  & Set5 & Set14 & B100 & Urban100 & Manga109 & \#Weighted-Avg\\
    \hline
    GRFormer-8$\times$32 & 800 &50.8 &32.58/0.8994& 28.88/0.7886& 27.75/0.7431& 26.90/0.8097& 31.31/0.9183& 28.80/0.8260\\
    GRFormer-16$\times$16& 800& 50.8& 32.53/0.8994& 28.89/0.7883& 27.75/0.7431& 26.92/0.8102& 31.34/0.9187& 28.81/0.8262\\
 \bottomrule
  \end{tabular}
}
\vspace{-1mm}
\end{table*}

\begin{table}[!t]
  \caption{Classical SR model comparisons ($\times$4 SR). We report PSNR on Urban100 and Manga109. All models are trained on DF2K training set.}
  \vspace{-3mm}
    \centering
  \label{tab: LargeModelComparison}
\footnotesize
\resizebox{0.48\textwidth}{!}{
 \centering
 \begin{tabular}{c|c|c|c|c|c}
    \toprule
     Method & window size& \#Params(/M) & \#MACs(/G)  & Urban100 & Manga109\\
    \hline
    SwinIR& 8$\times$8& 11.9& 747& 27.45& 32.03\\
    SRFormer& 16$\times$16& 10.52& 683& 27.53& 32.07\\
    GRFormer& 16$\times$16& 9.5& 641& 27.7& 32.16\\
 \bottomrule
  \end{tabular}
}
\vspace{-1mm}
\end{table}

\begin{table}[!t]
  \caption{ Evaluations of running time (/ms) on NVIDIA GeForce RTX 3080 GPUs.}
  \vspace{-3mm}
    \centering
  \label{tab: RunningTime}
\footnotesize
\resizebox{0.28\textwidth}{!}{
 \centering
 \begin{tabular}{c|c|c|c}
    \toprule
     Method & $\times$2 & $\times$3 & $\times$4\\
    \hline
    SwinIR-light& 183& 125& 109\\
    SwinIR-NG& 248& 185& 160\\
    GRFormer-light& 205& 167& 142\\
 \bottomrule
  \end{tabular}
}
\vspace{-4mm}
\end{table}

\noindent\textbf{Implementation Details.}
We set the GRSAB Group number, GRSAB number of a GRSAB Group, feature number, and attention head number, window size to 4, 6, 60, 3, 8$\times$32, respectively. The training low-resolution patch size we use is 64$\times$64. When training, we randomly rotate the images by $0^\circ$, $90^\circ$, $180^\circ$, $270^\circ$ and randomly flip images horizontally for data augmentation. We adopt the Adam optimizer with $\beta_1$ = 0.9 and $\beta_2$ = 0.99 to train the model for 600k iterations. The learning rate is initialized as $2 \times 10^{-4}$ and halves on \{250000, 400000, 510000, 540000\}-th iterations. We use L1 loss to train the model. The whole process is implemented by Pytorch on NVIDIA GeForce RTX 3080 GPUs.

\subsection{Comparisons with State-of-the-arts}
We compare our GRFormer with commonly used lightweight SR models for upscaling factor $\times$2, $\times$3, $\times$4, including EDSR \cite{EDSR}, CARN\cite{CARN}, LatticeNet\cite{LatticeNet}, SwinIR\cite{SwinIR}, SwinIR-NG\cite{SwinIR-NG}, HPINet\cite{HPINet}, DLGSANet\cite{DLGSANet}, SRFormer\cite{SRFormer}. We compare the parameters, calculations as well as performance on five commonly used SR benchmark datasets (Set5, Set14, B100, Urban100, Manga109). The comparison results are grouped for $\times$2, $\times$3, $\times$4 upscaling factor.

\textbf{Quantitative Comparison}
Table \ref{tab:QuantitativeComparison} shows quantitative comparisons in terms of PSNR and SSIM on five benchmark datasets. As shown in Table \ref{tab:QuantitativeComparison}, our GRFormer achieves the best PSNR score and SSIM score for $\times$2, $\times$3, $\times$4 task on Set14, Urban100, Manga109 and weighted average of the five benchmark datasets, and the PSNR score and SSIM score achieved by our GRFormer is either quite close or superior to that of SOTA model on Set5 and B100. What's more, GRFormer outperforms SOTA model on Urban100 and Manga109 by a large margin. It is worth noting that, our GRFormer outperforms SwinIR-light by a maximum PSNR of 0.42dB and SOTA by a maximum PSNR of 0.23dB, which is a significant improvement for image SR. Furthermore, our GRFormer outperforms SOTA model by about 0.1dB on the weighted average of the five benchmark datasets for $\times$2, $\times$3, $\times$4 task. Although our GRFormer achieves great performance, the parameters and MACs of GRFormer is relatively low. As shown in Fig. \ref{fig:PSNR_Params_MACs_comparison} (a), compared with the self-attention of other transformer-based SR models, our GRSA has the smallest number of parameters and calculations.

\textbf{Qualitative Comparison}
We further show visual examples of common used methods under scaling factor $\times$4. As shown in Fig.\ref{fig:expriment_result_comparison_092}, we use three images reconstructed by EDSR, CARN, LatticeNet, SwinIR, SwinIR-NG, SRFormer, HPINet and GRFormer to make qualitative comparisons. 

We qualitatively compare models on Urban100-img062, img028, and img074. For img062, EDSR, CARN, LatticeNet, SwinIR, SwinIR-NG, and SRFormer show severe texture and color distortions, especially in the lower right corner, while GRFormer has the smallest distortion and closest color to HR. For img028, models like EDSR, CARN, LatticeNet, and SRFormer exhibit significant blurring and deformation, whereas GRFormer achieves the best reconstruction. For img074, only GRFormer accurately reconstructs the clear window frame direction without distortion, highlighting its superior detail preservation.

\subsection{Ablation Analysis}
We conduct ablation experiments to study the effect of Grouped Residual Layer (GRL) and Exponential-Space Relative Position Bias (ES-RPB). Ablation experiments are trained on DF2K and evaluated on the Set5, Set14, B100, Urban100, Manga109 datasets. PSNR and SSIM are adopted to evaluate the perceptual quality of recovered images. We also adopt parameters and MACs on images with an upscaled spatial resolution of 1280 $\times$ 720 pixels to evaluate the complexity.
The results are shown in Table \ref{tab: AblationAnalysis}.

\textbf{Effectiveness of GRL}
As shown in Table \ref{tab: AblationAnalysis}, model $\textcircled{1}$ achieves a 47\% reduction in \#SA-Params, a 17\% reduction in \#Params, a 43\% reduction in \#SA-MACs, and a 17\% reduction in \#MACs compared to model $\textcircled{3}$, primarily due to optimizations in self-attention. Despite these reductions, GRL maintains performance, demonstrating its effectiveness.
Further experiments with models $\textcircled{4}$ and $\textcircled{5}$ reveal:
\begin{itemize}[leftmargin=*]
    \item Effectiveness of grouping scheme. Model $\textcircled{1}$ significantly reduces parameters and computations compared to model $\textcircled{4}$, with minor performance degradation mitigated by the residual structure in the QKV layer.

    \item Effectiveness of residual structure. Model $\textcircled{4}$, retaining the same parameters as model $\textcircled{3}$, significantly improves performance, especially on Urban100 and Manga109, by 0.07dB and 0.08dB, respectively.

\end{itemize} 

\textbf{Effectiveness of ES-RPB}
The core feature of ES-RPB is the ability to represent the pixel position information. To highlight the contribution of our ES-RPB, we replace the ES-RPB of model $\textcircled{1}$ with RPB in SwinIR to get model $\textcircled{2}$. As shown in Table \ref{tab: AblationAnalysis}, compared with model $\textcircled{2}$, model $\textcircled{1}$ improves the performance on five benchmark datasets in a large margin. Specifically, model $\textcircled{1}$ outperforms model $\textcircled{2}$ on Urban100 by 0.44dB PSNR score, which is a notable boost in lightweight image super-resolution. To further understand the reason of improvement brought by ES-RPB, we draw the three dimensional view of both ES-RPB in model $\textcircled{1}$ and RPB in model $\textcircled{2}$. As shown in Fig.\ref{fig:RPB_comparison}, we can see that, the curve of ES-RPB in model $\textcircled{1}$ is roughly similar to that of RPB in model $\textcircled{2}$, because both ES-RPB and RPB represent the relative position information. As shown in Fig.\ref{fig:RPB_comparison}, after comparing the curves of RPB and ES-RPB, we can find two differences: First, minor fluctuations on the curve of ES-RPB are much less than that of RPB, which means that most of noise interference is removed. Secondly, we can see from the curve of RPB that RPB is overfitted, which will lead to poor generalization ability. In contrast, our ES-RPB use simpler structure to improve generalization ability. It means that RPB learns a lot of content that is not universally applicable, which affects its generalization ability to represent relative position information.

\section{Comparison of window size}
As shown in Table \ref{tab: WindowSizeComparison}, GRFormer-$8\times32$ outperforms GRFormer-$16\times16$ by 0.05dB in Set5, while GRFormer-16$\times$16 outperforms GRFormer-8$\times$32 by 0.03dB in Manga109. Additionally, compared with Table~\ref{tab:QuantitativeComparison}, our GRFormer-16$\times$16 still achieves state-of-the-art results. Therefore, the choice between window sizes 8$\times$32 and 16$\times$16 does not significantly impact the model's overall performance.

\section{Classical Image Super-Resolution}
For the classical image SR task, we compare our GRFormer with two state-of-the-art Transformer-based methods: SwinIR, and SRFormer. Our GRFormer, SRFormer, and SwinIR are all 10M-scale models. As shown in Table \ref{tab: LargeModelComparison}, for $\times4$ task, our GRFormer outperforms SwinIR by 0.25dB and outperforms SRFormer by 0.17dB in Urban100. Therefore, our GRFormer still still outperforms the state-of-the-art Transformer-based methods in classical Image Super-Resolution.

\section{Running time}

As shown in Table \ref{tab: RunningTime}, the running times for the SwinIR-light, SwinIR-NG, and GRFormer-light methods are 109 ms, 160 ms, and 142 ms, respectively, for the $\times$4 task. These times were measured using images with an upscaled spatial resolution of 256$\times$256 pixels on NVIDIA GeForce RTX 3080 GPUs. Although GRFormer-light exhibits a slightly higher running time compared to SwinIR-light, it still offers a competitive performance with significantly lower running times than SwinIR-NG. Furthermore, GRFormer-light strikes a balance between efficiency and effectiveness, ensuring satisfactory performance while maintaining manageable computation times, thus making it a viable option for practical applications where both speed and quality are critical.

\section{Conclusion}
In this paper, we propose GRSA, an efficient self-attention mechanism composed of two components: GRL, which significantly reduces parameters and calculations, and ES-RPB, which efficiently and effectively represents distance-sensitive relative position information. The GRL module employs a grouping scheme to minimize redundancy in parameters and calculations with minimal performance loss and uses a residual structure to enhance feature learning for the QKV linear layer. Based on GRSA, we introduce GRFormer, a simple yet effective model for lightweight single image super-resolution. GRFormer leverages GRL and ES-RPB to significantly reduce parameters and MACs while enhancing PSNR and SSIM performance. Experimental results demonstrate GRFormer’s superior performance over previous state-of-the-art lightweight SR models on benchmark datasets, particularly Urban100 and Manga109. We hope GRSA will be a valuable tool for future SR model design research.

\bibliographystyle{ACM-Reference-Format}
\balance
\bibliography{sample-base}


\begin{thebibliography}{18}


\ifx \showCODEN    \undefined \def \showCODEN     #1{\unskip}     \fi
\ifx \showDOI      \undefined \def \showDOI       #1{#1}\fi
\ifx \showISBNx    \undefined \def \showISBNx     #1{\unskip}     \fi
\ifx \showISBNxiii \undefined \def \showISBNxiii  #1{\unskip}     \fi
\ifx \showISSN     \undefined \def \showISSN      #1{\unskip}     \fi
\ifx \showLCCN     \undefined \def \showLCCN      #1{\unskip}     \fi
\ifx \shownote     \undefined \def \shownote      #1{#1}          \fi
\ifx \showarticletitle \undefined \def \showarticletitle #1{#1}   \fi
\ifx \showURL      \undefined \def \showURL       {\relax}        \fi
\providecommand\bibfield[2]{#2}
\providecommand\bibinfo[2]{#2}
\providecommand\natexlab[1]{#1}
\providecommand\showeprint[2][]{arXiv:#2}

\bibitem[Ahn et~al\mbox{.}(2018)]%
        {CARN}
\bibfield{author}{\bibinfo{person}{Namhyuk Ahn}, \bibinfo{person}{Byungkon
  Kang}, {and} \bibinfo{person}{Kyung-Ah Sohn}.}
  \bibinfo{year}{2018}\natexlab{}.
\newblock \showarticletitle{Fast, accurate, and lightweight super-resolution
  with cascading residual network}. In \bibinfo{booktitle}{\emph{Proceedings of
  the European conference on computer vision (ECCV)}}.
  \bibinfo{pages}{252--268}.
\newblock


\bibitem[Bian et~al\mbox{.}(2021)]%
        {bian2021attention}
\bibfield{author}{\bibinfo{person}{Yuchen Bian}, \bibinfo{person}{Jiaji Huang},
  \bibinfo{person}{Xingyu Cai}, \bibinfo{person}{Jiahong Yuan}, {and}
  \bibinfo{person}{Kenneth Church}.} \bibinfo{year}{2021}\natexlab{}.
\newblock \showarticletitle{On attention redundancy: A comprehensive study}. In
  \bibinfo{booktitle}{\emph{Proceedings of the 2021 conference of the north
  american chapter of the association for computational linguistics: human
  language technologies}}. \bibinfo{pages}{930--945}.
\newblock


\bibitem[Choi et~al\mbox{.}(2023)]%
        {SwinIR-NG}
\bibfield{author}{\bibinfo{person}{Haram Choi}, \bibinfo{person}{Jeongmin Lee},
  {and} \bibinfo{person}{Jihoon Yang}.} \bibinfo{year}{2023}\natexlab{}.
\newblock \showarticletitle{N-gram in swin transformers for efficient
  lightweight image super-resolution}. In \bibinfo{booktitle}{\emph{Proceedings
  of the IEEE/CVF conference on computer vision and pattern recognition}}.
  \bibinfo{pages}{2071--2081}.
\newblock


\bibitem[Dong et~al\mbox{.}(2014)]%
        {SRCNN}
\bibfield{author}{\bibinfo{person}{Chao Dong}, \bibinfo{person}{Chen~Change
  Loy}, \bibinfo{person}{Kaiming He}, {and} \bibinfo{person}{Xiaoou Tang}.}
  \bibinfo{year}{2014}\natexlab{}.
\newblock \showarticletitle{Learning a deep convolutional network for image
  super-resolution}. In \bibinfo{booktitle}{\emph{Computer Vision--ECCV 2014:
  13th European Conference, Zurich, Switzerland, September 6-12, 2014,
  Proceedings, Part IV 13}}. Springer, \bibinfo{pages}{184--199}.
\newblock


\bibitem[He et~al\mbox{.}(2016)]%
        {ResidualLearning}
\bibfield{author}{\bibinfo{person}{Kaiming He}, \bibinfo{person}{Xiangyu
  Zhang}, \bibinfo{person}{Shaoqing Ren}, {and} \bibinfo{person}{Jian Sun}.}
  \bibinfo{year}{2016}\natexlab{}.
\newblock \showarticletitle{Deep Residual Learning for Image Recognition}. In
  \bibinfo{booktitle}{\emph{Proceedings of the IEEE Conference on Computer
  Vision and Pattern Recognition (CVPR)}}.
\newblock


\bibitem[Huang et~al\mbox{.}(2015)]%
        {Urban100}
\bibfield{author}{\bibinfo{person}{Jia-Bin Huang}, \bibinfo{person}{Abhishek
  Singh}, {and} \bibinfo{person}{Narendra Ahuja}.}
  \bibinfo{year}{2015}\natexlab{}.
\newblock \showarticletitle{Single Image Super-Resolution From Transformed
  Self-Exemplars}. In \bibinfo{booktitle}{\emph{Proceedings of the IEEE
  Conference on Computer Vision and Pattern Recognition (CVPR)}}.
\newblock


\bibitem[Li et~al\mbox{.}(2023)]%
        {DLGSANet}
\bibfield{author}{\bibinfo{person}{Xiang Li}, \bibinfo{person}{Jiangxin Dong},
  \bibinfo{person}{Jinhui Tang}, {and} \bibinfo{person}{Jinshan Pan}.}
  \bibinfo{year}{2023}\natexlab{}.
\newblock \showarticletitle{DLGSANet: lightweight dynamic local and global
  self-attention networks for image super-resolution}. In
  \bibinfo{booktitle}{\emph{Proceedings of the IEEE/CVF International
  Conference on Computer Vision}}. \bibinfo{pages}{12792--12801}.
\newblock


\bibitem[Liang et~al\mbox{.}(2021)]%
        {SwinIR}
\bibfield{author}{\bibinfo{person}{Jingyun Liang}, \bibinfo{person}{Jiezhang
  Cao}, \bibinfo{person}{Guolei Sun}, \bibinfo{person}{Kai Zhang},
  \bibinfo{person}{Luc Van~Gool}, {and} \bibinfo{person}{Radu Timofte}.}
  \bibinfo{year}{2021}\natexlab{}.
\newblock \showarticletitle{Swinir: Image restoration using swin transformer}.
  In \bibinfo{booktitle}{\emph{Proceedings of the IEEE/CVF international
  conference on computer vision}}. \bibinfo{pages}{1833--1844}.
\newblock


\bibitem[Lim et~al\mbox{.}(2017)]%
        {EDSR}
\bibfield{author}{\bibinfo{person}{Bee Lim}, \bibinfo{person}{Sanghyun Son},
  \bibinfo{person}{Heewon Kim}, \bibinfo{person}{Seungjun Nah}, {and}
  \bibinfo{person}{Kyoung Mu~Lee}.} \bibinfo{year}{2017}\natexlab{}.
\newblock \showarticletitle{Enhanced deep residual networks for single image
  super-resolution}. In \bibinfo{booktitle}{\emph{Proceedings of the IEEE
  conference on computer vision and pattern recognition workshops}}.
  \bibinfo{pages}{136--144}.
\newblock


\bibitem[Liu et~al\mbox{.}(2023)]%
        {HPINet}
\bibfield{author}{\bibinfo{person}{Jie Liu}, \bibinfo{person}{Chao Chen},
  \bibinfo{person}{Jie Tang}, {and} \bibinfo{person}{Gangshan Wu}.}
  \bibinfo{year}{2023}\natexlab{}.
\newblock \showarticletitle{From coarse to fine: Hierarchical pixel integration
  for lightweight image super-resolution}. In
  \bibinfo{booktitle}{\emph{Proceedings of the AAAI Conference on Artificial
  Intelligence}}, Vol.~\bibinfo{volume}{37}. \bibinfo{pages}{1666--1674}.
\newblock


\bibitem[Liu et~al\mbox{.}(2022)]%
        {SwinIR-V2}
\bibfield{author}{\bibinfo{person}{Ze Liu}, \bibinfo{person}{Han Hu},
  \bibinfo{person}{Yutong Lin}, \bibinfo{person}{Zhuliang Yao},
  \bibinfo{person}{Zhenda Xie}, \bibinfo{person}{Yixuan Wei},
  \bibinfo{person}{Jia Ning}, \bibinfo{person}{Yue Cao}, \bibinfo{person}{Zheng
  Zhang}, \bibinfo{person}{Li Dong}, {et~al\mbox{.}}}
  \bibinfo{year}{2022}\natexlab{}.
\newblock \showarticletitle{Swin transformer v2: Scaling up capacity and
  resolution}. In \bibinfo{booktitle}{\emph{Proceedings of the IEEE/CVF
  conference on computer vision and pattern recognition}}.
  \bibinfo{pages}{12009--12019}.
\newblock


\bibitem[Liu et~al\mbox{.}(2021)]%
        {SwinTransformer}
\bibfield{author}{\bibinfo{person}{Ze Liu}, \bibinfo{person}{Yutong Lin},
  \bibinfo{person}{Yue Cao}, \bibinfo{person}{Han Hu}, \bibinfo{person}{Yixuan
  Wei}, \bibinfo{person}{Zheng Zhang}, \bibinfo{person}{Stephen Lin}, {and}
  \bibinfo{person}{Baining Guo}.} \bibinfo{year}{2021}\natexlab{}.
\newblock \showarticletitle{Swin transformer: Hierarchical vision transformer
  using shifted windows}. In \bibinfo{booktitle}{\emph{Proceedings of the
  IEEE/CVF international conference on computer vision}}.
  \bibinfo{pages}{10012--10022}.
\newblock


\bibitem[Luo et~al\mbox{.}(2020)]%
        {LatticeNet}
\bibfield{author}{\bibinfo{person}{Xiaotong Luo}, \bibinfo{person}{Yuan Xie},
  \bibinfo{person}{Yulun Zhang}, \bibinfo{person}{Yanyun Qu},
  \bibinfo{person}{Cuihua Li}, {and} \bibinfo{person}{Yun Fu}.}
  \bibinfo{year}{2020}\natexlab{}.
\newblock \showarticletitle{Latticenet: Towards lightweight image
  super-resolution with lattice block}. In \bibinfo{booktitle}{\emph{Computer
  Vision--ECCV 2020: 16th European Conference, Glasgow, UK, August 23--28,
  2020, Proceedings, Part XXII 16}}. Springer, \bibinfo{pages}{272--289}.
\newblock


\bibitem[Shi et~al\mbox{.}(2016)]%
        {PixelShuffle}
\bibfield{author}{\bibinfo{person}{Wenzhe Shi}, \bibinfo{person}{Jose
  Caballero}, \bibinfo{person}{Ferenc Husz{\'a}r}, \bibinfo{person}{Johannes
  Totz}, \bibinfo{person}{Andrew~P Aitken}, \bibinfo{person}{Rob Bishop},
  \bibinfo{person}{Daniel Rueckert}, {and} \bibinfo{person}{Zehan Wang}.}
  \bibinfo{year}{2016}\natexlab{}.
\newblock \showarticletitle{Real-time single image and video super-resolution
  using an efficient sub-pixel convolutional neural network}. In
  \bibinfo{booktitle}{\emph{Proceedings of the IEEE conference on computer
  vision and pattern recognition}}. \bibinfo{pages}{1874--1883}.
\newblock


\bibitem[Timofte et~al\mbox{.}(2017)]%
        {DIV2K}
\bibfield{author}{\bibinfo{person}{Radu Timofte}, \bibinfo{person}{Eirikur
  Agustsson}, \bibinfo{person}{Luc Van~Gool}, \bibinfo{person}{Ming-Hsuan
  Yang}, {and} \bibinfo{person}{Lei Zhang}.} \bibinfo{year}{2017}\natexlab{}.
\newblock \showarticletitle{Ntire 2017 challenge on single image
  super-resolution: Methods and results}. In
  \bibinfo{booktitle}{\emph{Proceedings of the IEEE conference on computer
  vision and pattern recognition workshops}}. \bibinfo{pages}{114--125}.
\newblock


\bibitem[Vaswani et~al\mbox{.}(2017)]%
        {Self-Attention}
\bibfield{author}{\bibinfo{person}{Ashish Vaswani}, \bibinfo{person}{Noam
  Shazeer}, \bibinfo{person}{Niki Parmar}, \bibinfo{person}{Jakob Uszkoreit},
  \bibinfo{person}{Llion Jones}, \bibinfo{person}{Aidan~N Gomez},
  \bibinfo{person}{\L~ukasz Kaiser}, {and} \bibinfo{person}{Illia Polosukhin}.}
  \bibinfo{year}{2017}\natexlab{}.
\newblock \showarticletitle{Attention is All you Need}. In
  \bibinfo{booktitle}{\emph{Advances in Neural Information Processing
  Systems}}, \bibfield{editor}{\bibinfo{person}{I.~Guyon},
  \bibinfo{person}{U.~Von Luxburg}, \bibinfo{person}{S.~Bengio},
  \bibinfo{person}{H.~Wallach}, \bibinfo{person}{R.~Fergus},
  \bibinfo{person}{S.~Vishwanathan}, {and} \bibinfo{person}{R.~Garnett}}
  (Eds.), Vol.~\bibinfo{volume}{30}. \bibinfo{publisher}{Curran Associates,
  Inc.}
\newblock
\urldef\tempurl%
\url{https://proceedings.neurips.cc/paper_files/paper/2017/file/3f5ee243547dee91fbd053c1c4a845aa-Paper.pdf}
\showURL{%
\tempurl}


\bibitem[Zhang et~al\mbox{.}(2022)]%
        {zhang2022efficient}
\bibfield{author}{\bibinfo{person}{Xindong Zhang}, \bibinfo{person}{Hui Zeng},
  \bibinfo{person}{Shi Guo}, {and} \bibinfo{person}{Lei Zhang}.}
  \bibinfo{year}{2022}\natexlab{}.
\newblock \showarticletitle{Efficient long-range attention network for image
  super-resolution}. In \bibinfo{booktitle}{\emph{European conference on
  computer vision}}. Springer, \bibinfo{pages}{649--667}.
\newblock


\bibitem[Zhou et~al\mbox{.}(2023)]%
        {SRFormer}
\bibfield{author}{\bibinfo{person}{Yupeng Zhou}, \bibinfo{person}{Zhen Li},
  \bibinfo{person}{Chun-Le Guo}, \bibinfo{person}{Song Bai},
  \bibinfo{person}{Ming-Ming Cheng}, {and} \bibinfo{person}{Qibin Hou}.}
  \bibinfo{year}{2023}\natexlab{}.
\newblock \showarticletitle{Srformer: Permuted self-attention for single image
  super-resolution}. In \bibinfo{booktitle}{\emph{Proceedings of the IEEE/CVF
  International Conference on Computer Vision}}. \bibinfo{pages}{12780--12791}.
\newblock


\end{thebibliography}
\end{document}